\documentclass[10pt,journal]{IEEEtran}
\usepackage{hyperref}
\usepackage{graphicx}
\usepackage{float}
\usepackage{afterpage}
\usepackage{gensymb}
\usepackage{makecell}
\usepackage[utf8]{inputenc}

\ifCLASSINFOpdf
\else
\fi
\usepackage{amsmath}
\usepackage{upgreek}
\usepackage{titlesec}
\titlespacing*{\subsection}{1pt}{1pt}{1pt}
\titlespacing*{\section}{2pt}{2pt}{2pt}
\usepackage{cite}

\begin{document}
%

\title{3D Steering and Localization in Pipes and Burrows \\Using an Externally Steered Soft Growing Robot}
%
%
%
%
\author{Yimeng Qin,
        Jared Grinberg,
        William Heap, and~Allison~M.~Okamura,~\IEEEmembership{Fellow,~IEEE}
        
\thanks{This work was supported by the U.S. Department of Energy under Grant No. DE-AC02-05CH11231 through Lawrence Berkeley National Laboratory (LBNL), a donation from LAM Research Corporation, and the National Science Foundation under Grant No. 2345769.}
\thanks{Yimeng Qin, William Heap, Allison M. Okamura are with the Department of Mechanical Engineering, Stanford University, Stanford, CA 94305 USA \{yimengq, wheap, and
aokamura\}@stanford.edu. Jared Grinberg was with Stanford University during the research and is now with the Department of Robotics, University of Michigan, Ann Arbor, MI 4810 USA. }}

\maketitle

\IEEEtitleabstractindextext{%
\begin{abstract}
Navigation and inspection in confined environments, such as tunnels and pipes, pose significant challenges for existing robots due to limitations in maneuverability and adaptability to varying geometries. Vine robots, which are soft growing continuum robots that extend their length through soft material eversion at their tip, offer unique advantages due to their ability to navigate tight spaces, adapt to complex paths, and minimize friction. However, existing vine robot designs struggle with navigation in manmade and natural passageways, with branches and sharp 3D turns. In this letter, we introduce a steerable vine robot specifically designed for pipe and burrow environments. The robot features a simple tubular body and an external tip mount that steers the vine robot in three degrees of freedom by changing the growth direction and, when necessary, bracing against the wall of the pipe or burrow. Our external tip steering approach enables: (1) active branch selection in 3D space with a maximum steerable angle of 51.7°, (2) navigation of pipe networks with radii as small as 2.5 cm, (3) a compliant tip enabling navigation of sharp turns, and (4) real-time 3D localization in GPS-denied environments using tip-mounted sensors and continuum body odometry. We describe the forward kinematics, characterize steerability,  and demonstrate the system in a 3D pipe system as well as a natural animal burrow.

\end{abstract}

}

\maketitle


%
\IEEEpeerreviewmaketitle


%
%
%
%
\ifCLASSOPTIONcompsoc
\IEEEraisesectionheading{\section{Introduction}\label{sec:introduction}}
\else
\section{Introduction}
\label{sec:introduction}
\fi


Navigation and inspection in confined, tortuous, and tubular spaces, such as tunnels and pipes, are important to maintain civil infrastructure and ensure the safety of users and customers. Pipe systems are innumerous and critical for the transportation of essential resources such as freshwater, wastewater, oil, and gas. Worldwide, billions of dollars are spent annually to inspect, repair, and replace pipe systems, and hundreds of injuries and deaths occur each year due to pipe damage \cite{ASCE.}\cite{WaterBreak}\cite{PipeInjuries}. Gathering data from the interior of pipe systems is needed to detect pipe damage and determine when a pipe needs to be repaired or replaced. However, the interior of pipe systems is often inaccessible to humans and impermeable to electromagnetic signals, thus requiring mechanical systems to access and inspect them. Although external scanning methods such as ultrasonic or radiographic inspection are sometimes used to look for pipe damage, their effectiveness is limited when pipes are buried or surrounded by dense infrastructure or equipment. Moreover, these external methods cannot access or detect internal features such as internal corrosion, blockages, or cracks initiated from the inside.


\begin{figure}[t]
    \centering
    \includegraphics[width=0.5
    \textwidth]{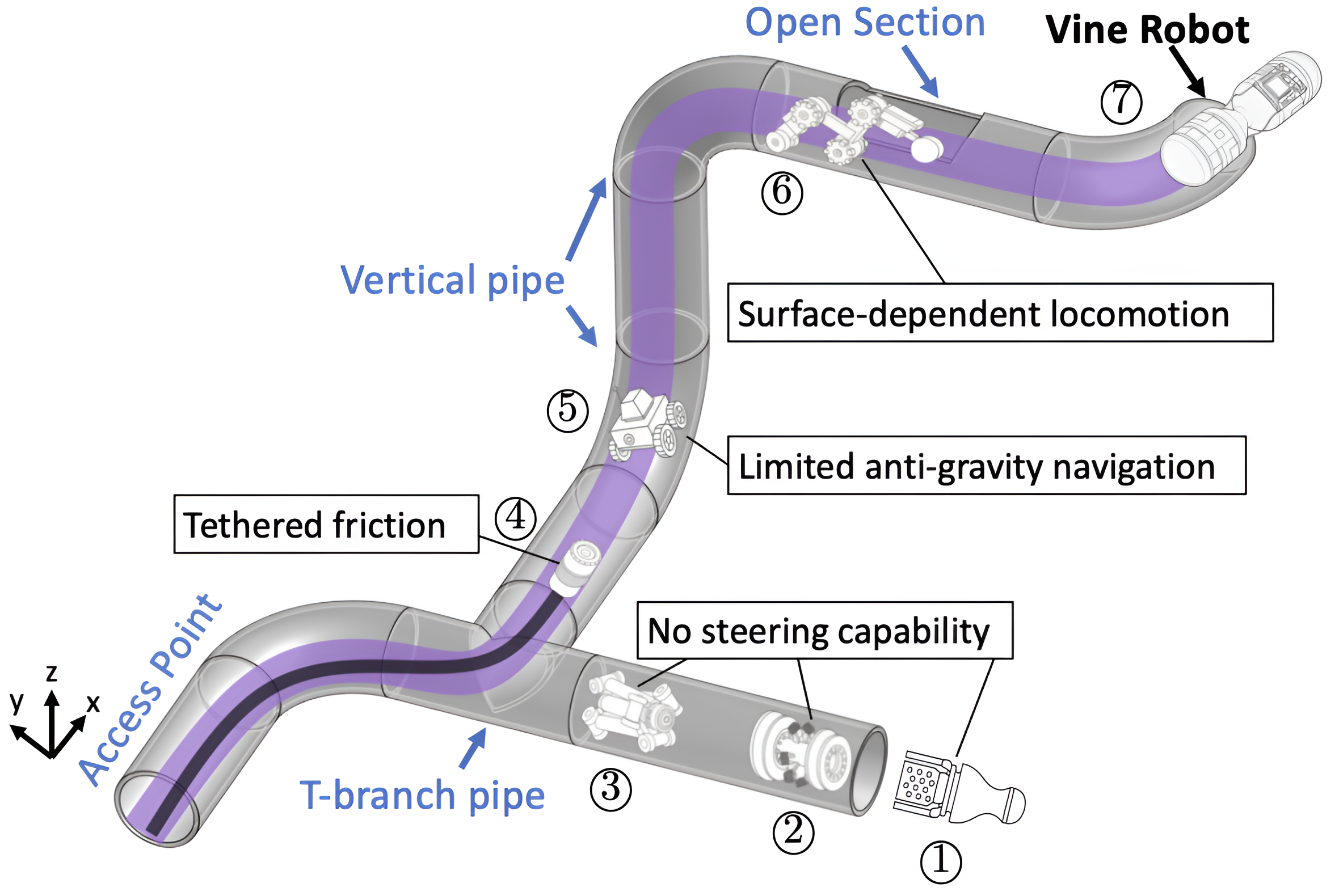}
    \caption{Conceptual illustration of a complex pipe system showcasing various existing types of in-pipe inspection robots, including floating robot \cite{pipemit}, smart PIGs \cite{PIG}, crawler/screw robots \cite{Vinten}, robotic scopes \cite{AIT_RiFlexio}\cite{Borescope}\cite{Waygate_MentorVisual}, wheeled/tracked robots \cite{Zhao2019}, worm/snake robots \cite{WormSnake}, and vine robots. The robot types \textcircled{1}-\textcircled{6} cannot navigate beyond the location shown in the drawing. (The conceptual representations of the all robots were generated using AI tools.)}
    \label{fig:pipeinspection}
\end{figure}

\begin{table}[ht]
\centering
\scriptsize
\caption{Comparison of Existing In-Pipe Inspection Robots}
\label{tab:robot_comparison}
\begin{tabular}{|c|p{1.8cm}|p{1.1cm}|p{1.1cm}|p{1.1cm}|p{1.1cm}|}
\hline
\textbf{\#} & \makecell{\textbf{Robot} \textbf{Type}} & \makecell{\textbf{Propul-}\\\textbf{sion}} & \makecell{\textbf{Agility}} & \makecell{\textbf{Adapt-}\\\textbf{ability}} & \makecell{\textbf{Anti-}\\\textbf{Gravity}} \\
\hline
\textcircled{1} & \makecell{Floating\\Robot \cite{pipemit}} & \makecell{Low} & \makecell{Low} & \makecell{High} & \makecell{Low} \\
\hline
\textcircled{2} & \makecell{Smart PIG} & \makecell{High} & \makecell{Moderate} & \makecell{Low} & \makecell{High} \\
\hline
\textcircled{3} & \makecell{Crawler/Screw\\Robot \cite{Ren2019DrivingReview}\cite{Yoon-GuKim2011DesignRobot}} & \makecell{Medium} & \makecell{Low} & \makecell{Medium} & \makecell{High} \\
\hline
\textcircled{4} & \makecell{Robotic\\Scope \cite{AIT_RiFlexio}\cite{Waygate_MentorVisual}} & \makecell{Low} & \makecell{High} & \makecell{High} & \makecell{Low} \\
\hline
\textcircled{5} & \makecell{Wheeled/Tracked\\Robot \cite{Zhao2019}} & \makecell{Medium} & \makecell{Low} & \makecell{Low} & \makecell{Low} \\
\hline
\textcircled{6} & \makecell{Worm/Snake\\Robot \cite{Virgala2020InvestigationPipe}\cite{pipeworm}} & \makecell{Medium} & \makecell{High} & \makecell{Medium} & \makecell{High} \\
\hline
\textcircled{7} & \makecell{Vine Robot } & \makecell{High} & \makecell{High} & \makecell{High} & \makecell{High} \\
\hline
\end{tabular}
\end{table}

\begin{figure*}[h]
    \centering
    \includegraphics[width=\textwidth]{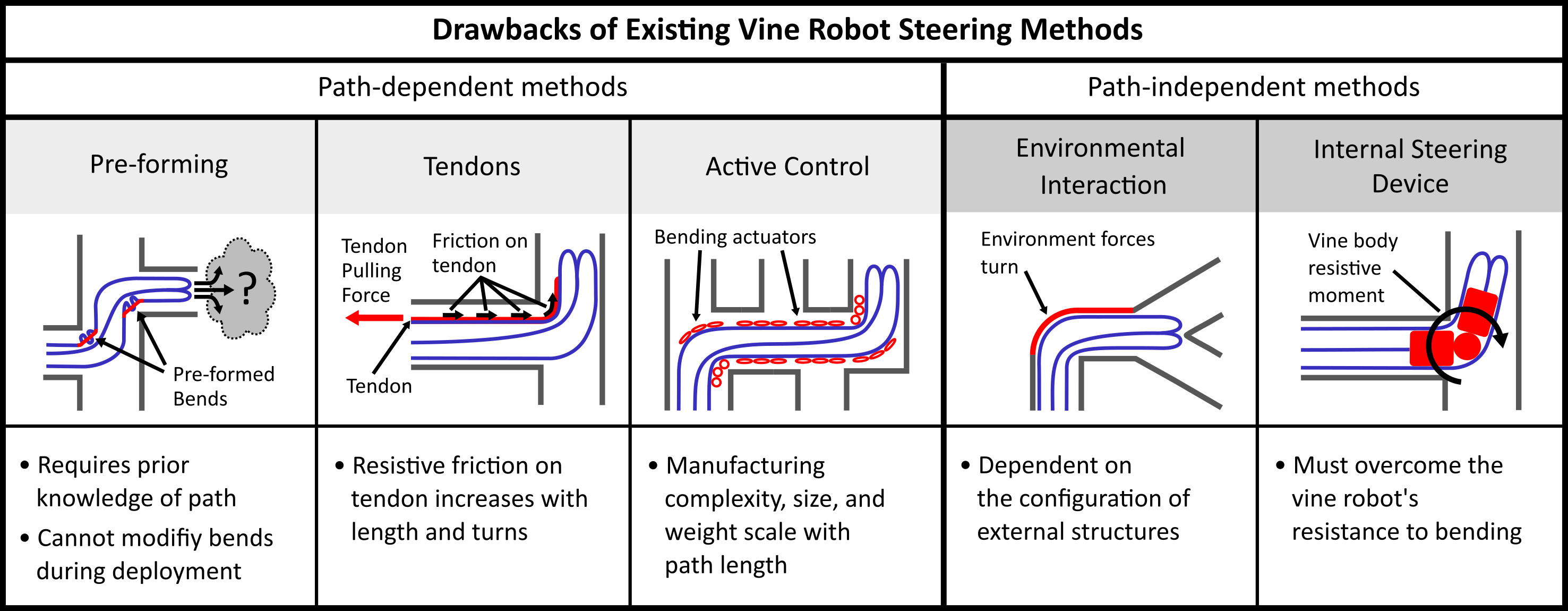}
    \caption{A comparison of different vine robot steering methods. While length-independent approaches address the limitations of length-dependent methods such as tendon-based steering and actuator-based active control, internal steering devices still face the challenge of overcoming the vine robot’s resistance to bending. In contrast, the external tip steering method we propose enables large, controlled curvature without deforming the entire body, offering a scalable and efficient solution for operation in confined environments.}
    \label{fig:steering_method}
\end{figure*}


As illustrated in Fig.~\ref{fig:pipeinspection}, while numerous mechanical and robotic systems have been developed and deployed for pipe inspection, they are still limited in their use cases. Commercially available in-pipe devices, such as pipeline inspection gauges or gadgets (PIGs), borescopes, rovers, and drones, are all limited by some combination of insufficient propulsion force, lack of ability to navigate pipes that branch, inability to travel within varying pipe diameters, and difficulty with traveling while opposing gravity \cite{PipeRobot}. To address these limitations, various novel robots have been developed to perform in-pipe inspections \cite{PipeReview}. Those robot designs can be categorized based on the propulsion mechanism as robotic scopes \cite{AIT_RiFlexio}\cite{Waygate_MentorVisual}, wheeled/tracked robots \cite{Zhao2019}, crawler/screw robots \cite{Ren2019DrivingReview}\cite{Yoon-GuKim2011DesignRobot}, worm-like/snake robots \cite{Virgala2020InvestigationPipe}\cite{pipeworm}, and passive floating robots \cite{pipemit}. As summarized in Table~\ref{tab:robot_comparison}, the comparison of existing in-pipe inspection robots is based on four key performance metrics. Propulsion refers to a robot’s ability to actively generate motion within the pipe environment. Agility explains how effectively a robot can maneuver through complex geometries such as bends, branches, and narrow sections. Adaptability describes the robot’s capability to conform to or traverse pipes with varying diameters, materials, or surface conditions. Lastly, anti-gravity performance indicates a robot’s ability to move against gravity, such as climbing vertically or navigating inclined sections of pipe. 

Table~\ref{tab:robot_comparison} highlights the advantages of different robot designs. For example, worm-like/snake robots and robotic scopes can perform delicate branch selection using a high-degree-of-freedom whole-body design or highly maneuverable robot tip. Crawler/screw robots and worm-like/snake robots use a wall-pressing friction-based motion, in which robots firmly press against the pipe walls to navigate in anti-gravity directions. These robots also commonly use expanding structures to accommodate a range of different pipe diameters. However, these prior robot designs face trade-offs between maneuverability, adaptability, and mechanical complexity. High degree-of-freedom designs typically require bulky actuation systems and complex control, which can limit their scalability or deployment in long or narrow pipe systems. Friction-based motion strategies can generate wear, require high energy input, and struggle in slippery or irregular-walled environments. Furthermore, rigid-body designs lack the compliance needed to navigate sharp turns or diameter-changing pathways, increasing the risk of mechanical jamming or structural damage.\par

Soft robots, due to their lightweight material, flexibility, and adaptability, can perform minimally invasive inspection tasks in confined environments \cite{soft_robot}. Vine robots, a type of pneumatic soft growing continuum robot, transport material from a fixed base and evert it at the tip driven by internal fluid pressure \cite{Hawkes2017AGrowth}. This mechanism offers many advantages over other robot types for navigating tight spaces. The vine robot can extend itself into torturous paths reaching significant lengths; travel with large propulsion force but negligible friction with respect to the environment; squeeze into apertures fractionally smaller than the body diameter; and create a self-supported continuum body to transport tail material and sensors \cite{large}\cite{internal}. \par

To navigate complex pipe networks, the vine robot's growth direction must be controlled. Controlling growth direction enables obstacle avoidance, targeted location inspection, and branch selection. Various steering designs have been proposed in the literature. Based on the actuation type, we divide them into two classes and describe the details in Section II. Some designs integrate soft pneumatic actuators to achieve 3D bending curvature on the main body \cite{Coad2019VineExploration}\cite{Naclerio2020SimpleMuscle}\cite{Kubler2022ASteering}\cite{Abrar2021HighlyStructure}. However, those designs cannot achieve large local curvature in three-dimensional space. Other designs use rigid internal steering devices to precisely control the tip orientation, but these mechanisms can only operate in a two-dimensional plane \cite{Haggerty2021HybridCapabilities}\cite{Takahashi2021EversionFunction}. An additional shortcoming of prior steering mechanisms is that bulky, rigid actuators limit the robot's ability to maneuver and conform within narrow or irregular environments. Shape-locking mechanisms can configure tip orientation and body shape, but non-negligible friction due to their mechanical design prevents them from deploying to arbitrary lengths \cite{Hawkes2017AGrowth}\cite{Jitosho2023PassiveRobots}\cite{M.Selvaggio20202020France.}.

In this letter, we present a novel external steering device with three degrees of freedom (DOFs) for vine robots to enable navigation in pipe and burrow environments and address previous vine robot limitations. Compared to other existing vine robot steering methods (as shown in Fig.~\ref{fig:steering_method}), our robot demonstrates (1) active branch selection in 3D with high local curvature and (2) functionality in long pipe networks with small radii. We use a compliant tip design to minimize control complexity and power consumption when the steering is not needed. Additionally, we implement a localization method based on body odometry and tip orientation, enabling real-time 3D tracking in GPS-denied environments. We validated our system through demonstrations in 2D and 3D pipe networks constructed in the laboratory as well as a field deployment into burrows created by California ground squirrels and used as a habitat by the endangered California Tiger Salamander (CTS).\par 

\section{Prior Work on Vine Robot Steering Methods}

Numerous techniques have been developed to control the path of vine robots in both constrained environments and open spaces. We categorize steering methods based on their dependence on vine robot length, as this distinction reflects fundamental differences in scalability, control complexity, and efficiency in confined environments. Broadly, vine robot steering methods fall into two categories: (1) Length-dependent steering methods, where the design, weight, or complexity of the steering mechanism scales with the robot’s length or the number of desired turns, and (2) length-independent steering methods, where both the steering performance and the physical characteristics of the mechanism—such as size, weight, and complexity—remain constant regardless of the robot’s length.




\subsection{Length-Dependent Steering Methods}
Some length-dependent steering methods of this type define the final shape of the vine robot before the vine robot is deployed, known as pre-forming \cite{Hawkes2017AGrowth}\cite{Helical}\cite{agharese2023}\cite{greer_robust_2020}. Pre-forming is achieved by creating a vine robot whose body passively inflates into the desired bent shape rather than a straight tube. This has been done by heating a thermoplastic tube to re-form it into the shape of a mold with the desired bends, or by pinching together and constraining two points along the length of the vine robot body. This produces a radially asymmetric contraction in the vine robot's body, which causes the vine robot to turn in the direction of the contraction. While pre-forming is a simple method that can achieve complex curvatures, the shape of the vine robot body cannot be modified while the vine robot is deployed. 

Active control of vine robot body contraction was introduced to steer vine robots during deployment. One approach involves using latches or dissolvable string to selectively constrain the extra material between two pinched points of the vine robot body \cite{Hawkes2017AGrowth}\cite{Kubler2022ASteering}\cite{string}. Alternatively, soft pneumatic actuators mounted along the vine robot body can provide asymmetric strain \cite{greer_series_2017}\cite{kubler_comparison_2023}\cite{Abrar2021HighlyStructure}. While this method provides control of whether bending occurs and to what degree, soft pneumatic actuators are strain-limited, which leads to low local curvature. In contrast, tendon-based steering enables controlled high-curvature bending. This method routes tendons along the length of the vine robot body and generates local asymmetric contraction by pulling the tendons from the robot’s proximal end \cite{Geometric}\cite{ninatanta_design_2024}\cite{BrianReconfigurable}. Tendon steering is simple and lightweight, with only the mass of a small diameter rope or wire scaling with vine robot length. However, the friction between tendons and the vine robot body quickly makes it impractical to use in long, winding paths.

More broadly, these methods all produce bends whose position on the vine robot body cannot be modified during deployment of the vine robot. As a result, the number of modifications or actuators, and thus manufacturing complexity, the amount of material that must be transported through the pipe, and the size and weight of the vine robot all scale with the number of bends required. This introduces length scaling issues: as the number of bends increases, so does the system’s complexity. In many cases, desired bend locations are not known before deployment, so the vine robot's position relative to branches or obstacles cannot be precisely controlled. To prevent a lengthy trial and error process, the density of actuators or modified sections needs to be increased along the vine robot body to allow for a wider range of potential bend positions.

\subsection{Length-Independent Steering Methods}

To overcome the length scaling issues described above, the simplest method is bending through environmental interaction. For example, by having the vine robot grow into a wall at specific angles, the vine robot can be forced by the wall to turn left or right \cite{Obstacle-aid}. Environmental interaction has been used to direct a vine robot to different endpoints from the same starting location by only changing the vine robot's initial growth direction. This approach relies on the presence of external obstacles to steer the robot passively. In pipe environments, the walls and bends similarly serve as “obstacles”, which enables passive navigation without active steering mechanisms in some scenarios. This can be useful in a highly constrained environment like a pipe because the vine robot will passively follow bends in the pipe. However, environmental interaction relies on obstacles being present in useful locations, and does not allow the vine robot to select between branching paths. 

In addition to relying on the environment, the vine robot itself can control the path. For example, internal steering devices have been developed that are \textit{not} attached to the main tube body; these are placed inside the vine robot body and can move along the vine robot or stay at the tip \cite{Takahashi2021EversionFunction}. The device itself bends, such as by having two bodies pivot relative to each other, which creates a bend in the vine robot body. For navigation in highly confined environments such as pipes and burrows, only one device is needed, no matter the number of bends encountered. Additionally, because bending force is provided locally, frictional losses do not scale, unlike tendons. However, the two internal steering devices previously developed are limited in their size and turning ability. In \cite{Haggerty2021HybridCapabilities}, the internal steering device presented has a diameter of 70 mm and length of 215 mm (though it also provides additional reeling capabilities), and steers within a two-dimensional plane. In \cite{Maur2021RoBoa:Applications}, the internal steering device has a diameter slightly under 100 mm and a length over 600 mm, and uses pneumatic actuators that limit its achievable local bending curvature. Both devices are thus unable to travel in small-diameter pipes or burrows with tight, three-dimensional turns. Beyond the specific drawbacks of these designs, internal steering devices must also always overcome an inflated vine robot body's resistance to bending, an inherent challenge of the internal steering approach that necessitates stronger and bulkier components and actuators. 

Instead, for length-independent path selection in highly constrained environments such as pipes or burrows, we propose an external tip steering mechanism to actively select branches, while the body passively follows the path determined by the leading tip. This approach offers the scalability of internal steering devices while maintaining the efficiency and simplicity of environmental interaction.

\section{Design Concept}
\begin{figure}[!b]
    \centering
    \includegraphics[width=0.48
    \textwidth]{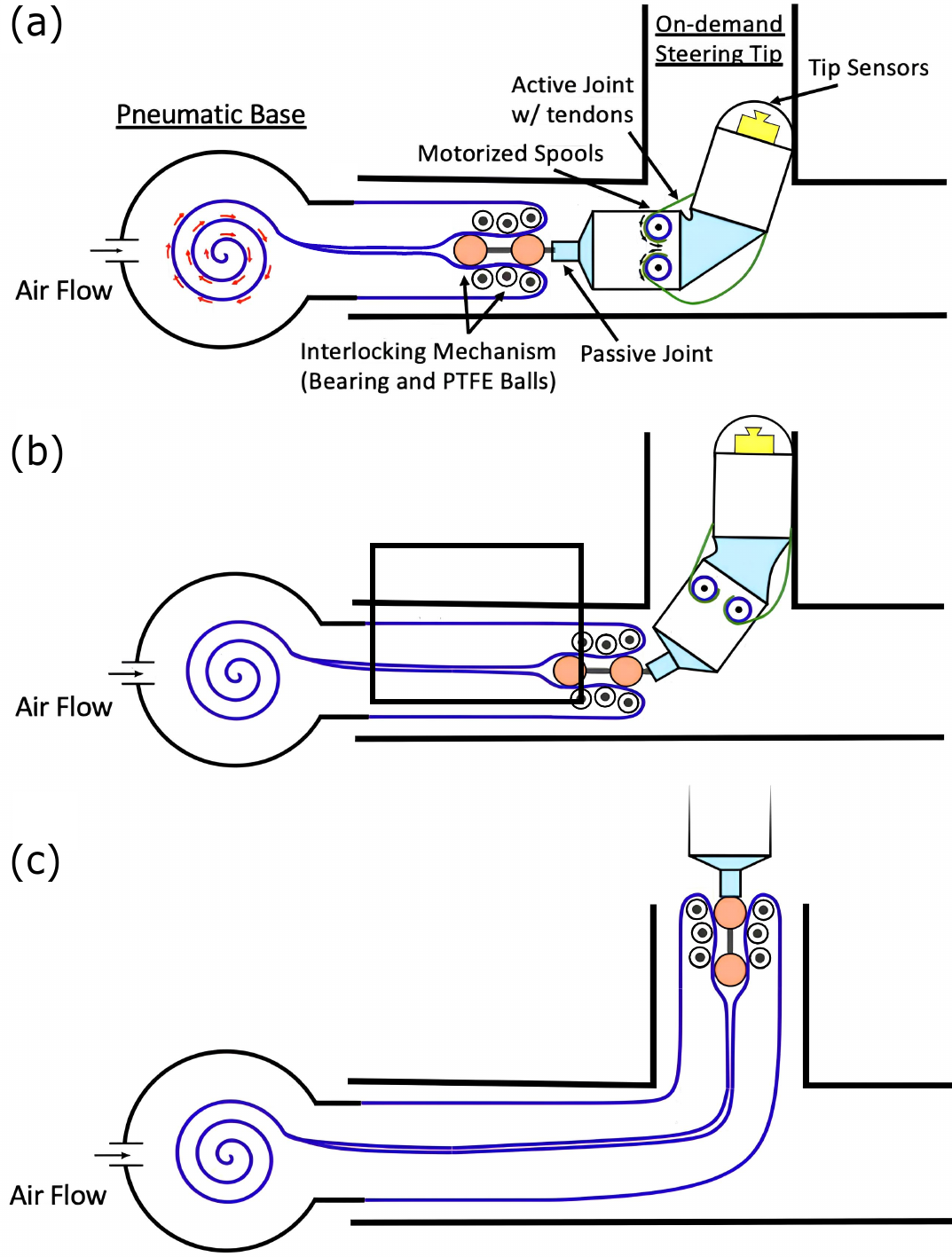}
    \caption{The design concept for integrating an external steering tip with an interlocking tip mount onto the vine robot to enable active navigation in branched pipe environments. The sequence illustrates three steps for steering into a sharp T-branch: (a) Activating the tendon-driven steering mechanism to guide the tip into the desired branch; (b) Extending the robot body, allowing the passive joint to conform to the environment; (c) Continuing the extension until the rigid tip section fully enters the target branch.}
    \label{fig:tipconcept}
\end{figure}

We use external tip steering to guide the soft robot's growth direction. Within a confined tubular environment, the pipe walls naturally constrain the robot body. Therefore, instead of re-configuring the whole vine robot body, the steering only needs to happen at the distal robot end. Due to the compliance of the soft robot body, or, infinite DOF, the tail material will feed through the body, following the previous path within the left-behind body. The high-level steering concept is illustrated in Fig.~\ref{fig:tipconcept}. The tip steering mechanism is first angled to direct the robot toward the desired branch. Once the tip is aligned, the vine robot extends forward, enabling the compliant tip section and the soft vine robot body to passively follow and navigate into the branch.

Relying on the above design principle, our robot system consists of an external steerable tip mechanism and a tip mount device. Because the tail material extends at twice the speed as the tip material relative to the ground \cite{Sadeghi2017TowardTechnologies}\cite{Tsukagoshi2011TipTerrain}, the tip mount design must ensure that the steering location occurs at the robot's tip. We build on a previous tip mount design that is passive, compact, and reliable \cite{internal}. The steering is achieved by incorporating a tendon-driven mechanism: a spherical joint is placed in the center of the device, surrounded by three parallel cables. By pulling each cable using a motorized spool, the tip section bends toward the corresponding direction. When steering against gravity, the actuation system must generate sufficient torque to lift the electronics and hardware components mounted at the tip. Additionally, to reduce the control complexity and power consumption, the tip section should have some flexibility to passively navigate through the tortuous path. To achieve this, soft spherical joints molded from silicone were designed with anisotropic stiffness—high axial stiffness to prevent structural collapse under compressive loads, and low bending stiffness to allow directional flexibility. The hollow inner design of the rigid sections is used to store all the electronic components (e.g., actuators, sensors, control units, etc). 

\section{Modeling}
Here, we describe the external steering device mathematically. This consists of the forward kinematics of the whole device and the silicone-molded soft joints and force production. The model is verified by experimental results in Section VI. \par 


\subsection{Forward Kinematics}
\begin{figure}[h]
    \centering
    \includegraphics[width=0.48
    \textwidth]{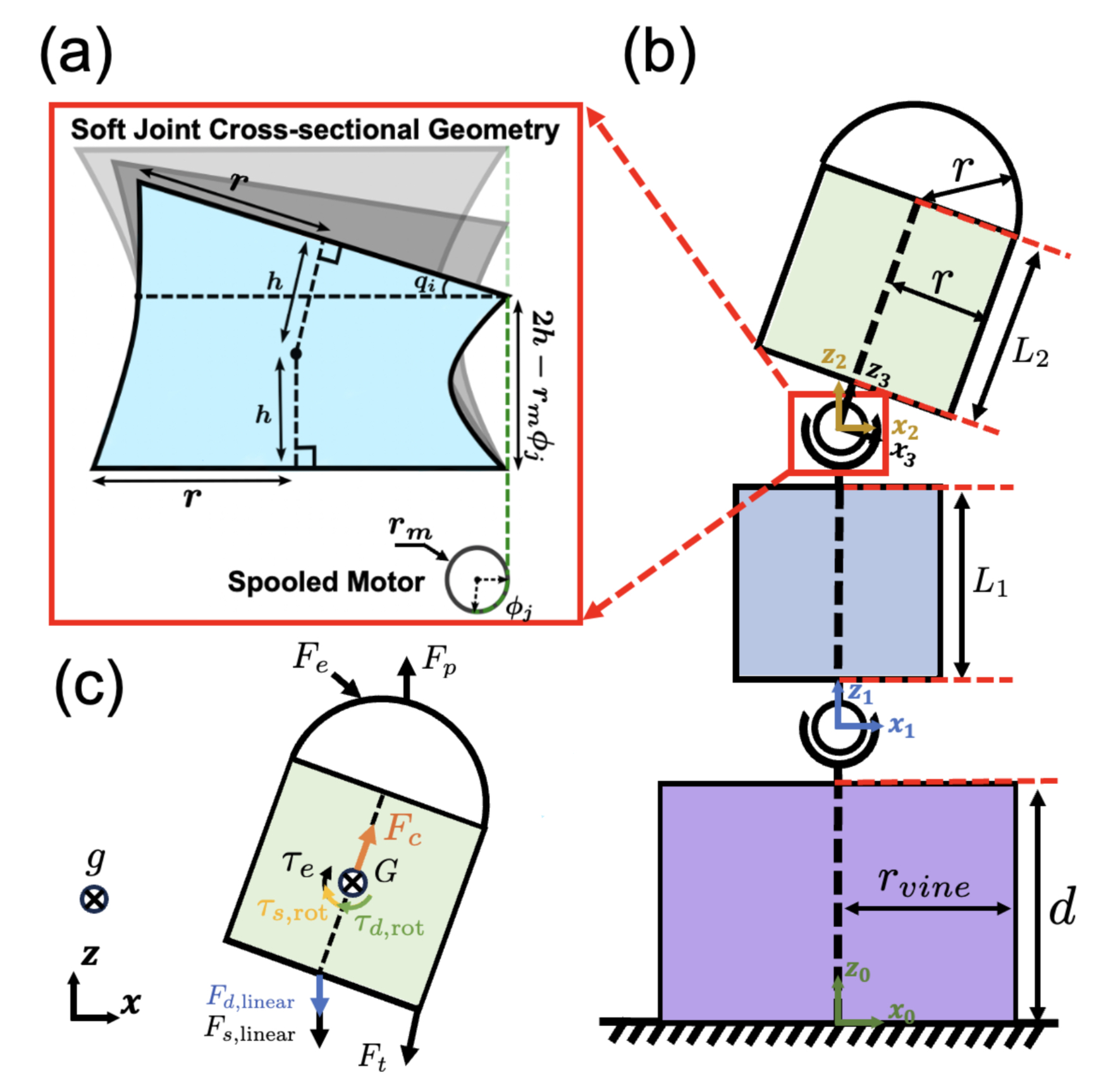}
    \caption{Forward kinematics modeling and free body diagram of the external tip steering mechanism. (a) Cross-sectional view of the deformed geometry of the silicone-based soft joint, including the tendon and spooled motor. (b) A simplified kinematic model for the mechanism using a spherical joint to represent the soft joint. (c) Free body diagram of the active (distal) joint.}
    \label{fig:kinematics}
\end{figure}

1) \textit{External Steering Device}: We model the external steering device as a simplified prismatic-spherical-spherical (PSS) robot. As shown in Fig.~\ref{fig:kinematics}(b), the vine robot body without environmental interaction behaves like a prismatic joint, and we assume the silicone-molded soft joint behaves like a spherical joint. The assumption is based on the origami-inspired geometry of the design, whereby the joint rotates without undergoing axial compression. 

With four standard reference frames defined in Fig.~\ref{fig:kinematics}(b), the 3D homogeneous transformation matrix from frame $\{i-1\}$ to frame $\{i\}$ is,
\begin{equation}
\label{eqn:trans}
{ }_i^{i-1} T=\left[\begin{array}{cc}
{ }_i^{i-1} R & ^{i-1} P_{\text {orig }} \\
0 & 1
\end{array}\right] \quad \text { with } i \in[1,3]
\end{equation}
where ${ }_i^{i-1} R$ is the corresponding rotation matrix and $^{i-1} P_{\text {orig }}$ is the position vector from frame $\{i-1\}$'s origin to frame $\{i\}$'s origin, expressed in frame $\{i-1\}$ frame.

Following the fixed angle representation convention, the rotation matrix can be expressed as,
\begin{equation}
\label{eqn:rot}
{ }_i^{i-1} R = R_{z}(\gamma_i)R_{y}(\beta_i)R_{x}(\theta_i) \\
\end{equation}
where $R_{z}(\gamma_i)$,$R_{y}(\beta_i)$ and $R_{x}(\theta_i)$ stand for single axis rotation matrix in z, y and x axis with angle $\gamma_i$, $\beta_i$ and $\theta_i$, respectively. Angle $\theta_i$, $\beta_i$, and $\gamma_i$ represent rotation angle of rotating frame$\{i-1\}$ clock-wise around x, y and z axis with respective to frame$\{i\}$. Based on the soft joint geometry shown in Fig.~\ref{fig:kinematics}(a), the maximum rotation angle happens when two rigid sections intersect with each other, which is $\theta_{i,max} = \alpha_{i,max}=52.5 \degree$.

Using Equations~(\ref{eqn:trans}) and (\ref{eqn:rot}), the robot tip location could be expressed in any chosen reference frame. Given the end effector position, $^{\{3\}} P_{t}$, in frame $\{3\}$, the relative position, $^{\{k\}} P_{t}$, in frame $\{k\}$ can be expressed as, 
\begin{equation}
\label{eqn:kinem}
{ }^{\{k-1\}} P_t=\left(\prod_{i=k}^{i=3}{ }_i^{i-1} T\right) \cdot{ }^{\{3\}} P_t \quad \text { with } k \in[1,3]
\end{equation}

\noindent The tip position in frame $\{3\}$ (end-effector frame) is
\begin{equation}
\nonumber
\label{eqn:tip_frame3}
^{\{3\}} P_t = 
\left[\begin{array}{cc}
0 \\
0 \\
r_{\mathit{eff}} \\
\end{array}\right] {,}
\end{equation}

\noindent The effective radius is \begin{equation}
    r_{\mathit{eff}} = h+L_{2}+r
\end{equation}
where $h$ is half height of soft joint, $r$ is the rigid cylinder container radius, and $L_{2}$ is the length of tip container.

To represent the tip position in frame$\{2\}$, $k=3$ is substituted into Equation~\ref{eqn:kinem},

\begin{equation}
^{\{2\}} P_t = r_{\mathit{eff}}
\left[\begin{array}{cc}
 \sin (\gamma_3) \sin (\theta_3)+ \cos (\gamma_3) \sin (\beta_3) \cos (\theta_3) \\
 \sin (\beta_3) \cos (\theta_3) \sin (\gamma_3)- \cos (\gamma_3) \sin (\theta_3) \\
 \cos (\beta_3) \cos (\theta_3) \\
\end{array}\right]
\end{equation} 

2) \textit{Soft Joint}: Here, we find an analytical expression to map each motor's rotation to the soft joint's angle ($i=3$). The detailed schematic is shown in Fig.~\ref{fig:kinematics}(a).

Three motors control rotation in the x-z plane (right and left) and rotation in the y-z plane (up), where we notate the motor rotation angles as $\phi_{j}$ with $j=0,1,2$, respectively. Assuming that the motion of the tendons does not generate a twisting rotation in the z-direction, only two soft joint angles, denoted by $\theta_{i}$ and $\beta_{i}$, with their respective signs, are sufficient to define the mapping. For notational simplicity, we introduce $q_{j}$ as the rotation angle responding to $j^{\mathit{th}}$ motor's rotation:

\begin{equation}
q_{j}=\left\{\begin{array}{lll}
\theta_3 < 0 & j=0 \\
\theta_3 > 0 & j=1 \\
\beta_3 & j=2 , 
\end{array}\right.
\end{equation}

According to the simplified soft joint geometry defined in Fig.~\ref{fig:kinematics}, the joint angle $q_j$ is expressed as,
\begin{equation}
q_{j}=(-1)^j\left[\tan^{-1}\left(\frac{h}{r}\right)-\sin ^{-1}\left(\frac{h-r_m \varphi_j}{\sqrt{h^2+r^2}}\right)\right]
\end{equation}
where $r_{m}$ is the radius of motor spool, and $\varphi_j$ represents the rotation angle of $j^{th}$ motor spool.

\subsection{Force Analysis}

To evaluate the efficacy of the tendon-driven actuation system, we determine the maximum load the tip can support before spool motor failure occurs. The system must be robust enough to maneuver the first segment of the tip in three dimensions, while overcoming the weight of the segment, the camera, the microcontroller, the IMU, and other electronic components. In this section, we first provide general dynamics modeling, followed by experimental results for the maximum weight the tip can move with a given motor. 

The force analysis is conducted on the robot's tip section with respect to frame $\{3\}$, as shown in Fig.~\ref{fig:kinematics}(c). The tip section experiences pulling force generated by a motorized spool and transmitted through a tendon, $F_t$; gravitational force due to the tip section’s mass, $G$; acceleration defined at center of mass, $\vec{a}$; propulsion force provided by the body $F_{p}$; external forces such as friction and normal force exerted by the environment, $F_{e}$; and centrifugal force due to non-inertial frame $\{3\}$, $F_{c}$. The soft joint model is a 6-DOF spring-damper system with 3-DOF linear translation and 3-DOF rotation. Thus, the tip section also experiences linear spring-damper forces, $F_{s,\text{linear}}$ and $F_{d,\text{linear}}$. The force balance equation is
\begin{equation}
m \vec{a}={G}+{F_t}+{F_e}+{F_p}+{F_c}+{F_{s,\text{linear}}}+{F_{d,\text{linear}}} .
\end{equation}
Specifically,
\begin{equation}
    \nonumber
    F_{\text {p}} = \frac{1}{2} P_{b}A_{b}
\end{equation}
\begin{equation}
    \nonumber
    F_{c} = \frac{1}{2} \left( \frac{L_2}{2} + h \right) \left( \dot{\theta}_3^2 + \dot{\beta}_3^2 + \dot{\gamma}_3^2 \right)
\end{equation}
\begin{equation}
\nonumber
F_{s,\text{linear}}
 =\left[\begin{array}{ccc}
k_x & 0 & 0 \\
0 & k_y & 0 \\
0 & 0 & k_z
\end{array}\right]\left[\begin{array}{l}
\Delta x \\
\Delta y \\
\Delta z
\end{array}\right]
\end{equation}
\begin{equation}
\nonumber
F_{d,\text{linear}}
 = \left[\begin{array}{ccc}
b_x & 0 & 0 \\
0 & b_y & 0 \\
0 & 0 & b_z
\end{array}\right]\left[\begin{array}{c}
\Delta \dot{x} \\
\Delta \dot{y} \\
\Delta \dot{z} \\
\end{array}\right] \raisebox{-3ex}{,}
\end{equation}
where $P_b$ and $A_b$ are the vine robot body's pressure and cross sectional area respectively; $\Delta x$, $\Delta y$, and $\Delta z$ are the soft joint deformation in the x, y, and z direction; and $k_x$, $k_y$, $k_z$, $b_x$, $b_y$, and $b_z$ are the corresponding spring constants and damping coefficients.

\begin{figure*}[!t]
  \centering
  \includegraphics[width=\textwidth]{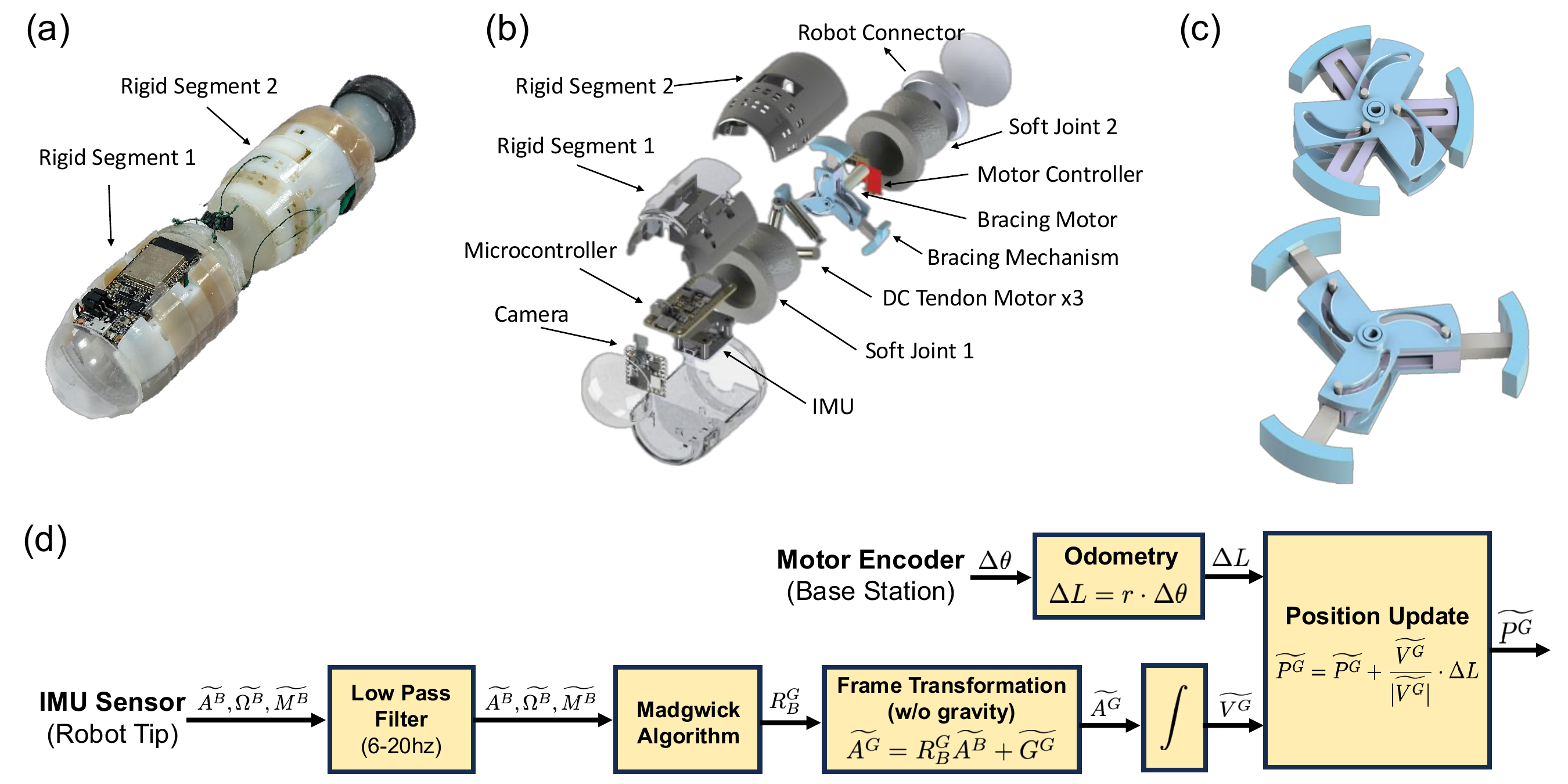}
  \caption{Implementation of the external tip steering device: (a) Image and (b) CAD rendering. The design includes a tendon-driven mechanism for in-pipe navigation, a bracing mechanism to enable operation across varying pipe diameters, and a sensing module that integrates localization sensors and inspection tools (IMU and camera). A zoomed-in view of the bracing mechanism is shown. The battery module is omitted for clarity. (c), showing the shrinking (top) and expanding (bottom) state. The software workflow diagram (d) depicts the sensor data processing and computation used to localize the robot tip position.}
  \label{fig:design_figure}
\end{figure*}

In addition to the translational forces defined above, the tip section also experiences torques resulting from those forces and the moment arms between the force application points and the tip's center of mass. The position vector (i.e., the moment arm) is defined from the frame $\{3\}$ origin to the location where the corresponding force is being applied. The position vectors associated with tendon force, gravitational force, and external force are $r_T$, $r_G$, and $r_e$, respectively. The tip section also experiences rotational spring/damper torque due to deformation of the soft joint, $\tau_{s,\text{rot}}$ and $\tau_{d,\text{rot}}$. The angular acceleration, $\vec{\alpha}$, is defined as the rotational acceleration acting at the center of mass. The reaction moment is $\tau_e$. The torque balance is,
\begin{equation}
I \vec{\alpha}=\tau_e + r_t \times F_t + r_G \times F_G + r_e \times F_e + {\tau_{\text {s,rot}}}+{\tau_{\text {d,rot}}}
\end{equation}
\begin{equation}
\nonumber
\tau_{s,\text{rot}}
 =\left[\begin{array}{ccc}
k_{\theta} & 0 & 0 \\
0 & k_{\beta} & 0 \\
0 & 0 & k_{\gamma}
\end{array}\right]\left[\begin{array}{l}
\Delta \theta_3 \\
\Delta \beta_3 \\
\Delta \gamma_3
\end{array}\right]
\end{equation}
\begin{equation}
\nonumber
\tau_{d,\text{rot}}
 = \left[\begin{array}{ccc}
b_{\theta} & 0 & 0 \\
0 & b_{\beta} & 0 \\
0 & 0 & b_{\gamma}
\end{array}\right]\left[\begin{array}{c}
\Delta \dot{\theta_3} \\
\Delta \dot{\beta_3} \\
\Delta \dot{\gamma_3}
\end{array}\right] \raisebox{-3ex}{,}
\end{equation}
where $\Delta \theta_3 $, $\Delta \beta_3 $,and $\Delta \gamma_3$ are the soft joint rotational deformations about the x, y, and z direction; $k_\theta$, $k_\beta$, $k_\gamma$, $b_\theta$, $b_\beta$, and $b_\gamma$ are the corresponding spring constants and damping coefficients.

To evaluate the maximum tip weight that the tendon-driven system can maneuver using a spooled motor with a given peak torque, we made the following assumptions. Because the silicone-molded joint has low stiffness, we assume that torque and force due to soft joint deformation are negligible compared with other force components. In other words, $F_{s,\text{linear}} = F_{d,\text{linear}} =\tau_{s,\text{rot}} =\tau_{d,\text{rot}} = 0$. For implementation, we use a high gear ratio spool motor (details in Section V-A) to achieve high tendon force. Therefore, we assume the soft joint rotates at a slow rotational speed with a high payload, which means $F_c \approx 0$. Furthermore, to enable precise steering, the vine robot maintains a slow-growing speed during steering, which means $F_p \approx 0$. Therefore,

\begin{equation}
\nonumber
\tau_{e} + r_t \times F_t + r_G \times G = r_e \times F_e
\end{equation}
\begin{equation}
\nonumber
F_{e} = G + F_t
\end{equation}
\begin{equation}
\label{eq:blockedforce}
\tau_m = r_m \times F_t,
\end{equation}
where $\tau_m$ is the torque provided by the spool motor. 
\section{Design Implementation}
\subsection{Mechanical Design}

1) \textit{Steering Mechanism}: As shown in Figs.~\ref{fig:design_figure}(a) and \ref{fig:design_figure}(b), the design integrates a suite of components for steering: a microcontroller (MCU) for electrical coordination, a camera for directional guidance, an Inertial Measurement Unit (IMU) for positional tracking, an actuation system for movement, stability features for 3D navigation, and a battery for power supply. The electrical components will be further explained in Section V-B. The constraints imposed by the physical environment necessitated a compact design, leading to the creation of a tip mount comprising two rigid segments connected by silicone-molded flexible soft joints (Ecoflex™ 00-50). These soft joints facilitate bending and are designed with internal channels for electrical wiring, connecting the various components housed in the rigid segments.

The second rigid segment (Rigid Segment 2, Fig.~\ref{fig:design_figure}(a)) incorporates a steering mechanism. This system uses three motors (Pololu 700:1 Plastic Planetary Gearmotor) attached to custom 3D-printed spools, which manipulate tendons attached to the first rigid segment. This arrangement enables controlled bending in multiple directions—left, right, and upwards into three-dimensional space. Downward movement is left to passive mechanisms, primarily gravity.

As the motors rotate, the tendons begin to wind around the spools, increasing the tension in the tendons. The tendons are routed through miniature holes in the rigid body of the device, and then attached to the first rigid segment for actuation. The directional actuation of the tip of the steering device is caused by the tension in the tendons.

2)\ \textit{Bracing Mechanism}: A key challenge in 3D navigation is maintaining rotational stability about the axis of the vine robot body. Inspired by \cite{grabcad}\cite{Nikhade2024}, we designed a bracing mechanism that converts rotation into linear motion, located within the second rigid segment. As shown in Fig.~\ref{fig:design_figure}(c), this mechanism, when activated, extends three stabilizing legs outward to the pipe walls, providing the necessary support and friction to prevent rotational movement. This stability is crucial when the robot's first segment is raised to navigate upwards or in a non-horizontal plane. Upon completion of the 3D movement, the bracing mechanism retracts, allowing the robot to progress smoothly. In its retracted state, the mechanism has the same outside diameter as the second rigid segment of the tip steering mechanism, fitting into it seamlessly. In its extended state, the mechanism has an outside diameter of 61 mm.

\subsection{Electronics and Software}
The electronic architecture of our design is anchored by two ESP32-based microcontroller units (MCUs). These MCUs communicate wirelessly, ensuring seamless teleoperation and sensor data communication. One MCU interfaces at the user side to receive real-time robot tip orientation, communicate with the motor that releases or stores vine robot material to control growth, and transmit teleoperation steering commands based on user keyboard input. Each command specifies the motor number, PWM value, and actuation duration. For example, the command 1,100,10 informs motor 1 to actuate at 100 duty cycle for 10 seconds. Based on the teleoperation command, the second MCU embedded within the first/second rigid section of the steering mechanism will control the spool motors via a customized motor controller board with a DRV8838 IC chip. A 9-axis Inertial Measurement Unit (IMU - Adafruit BNO055) and a compact WiFi camera (Arduino Vision Nicla) are also integrated at the tip section to provide tip orientation and vision. Furthermore, a 3.7V Lithium Polymer (LiPo) battery with a power regulation system is placed at the second rigid section to provide onboard power.\par


As the vine robot grows via continuous eversion of new body material, we estimate the tip odometry $\Delta L$ by tracking the length of material released over time. The uneverted body material is stored on a spool, as described in \cite{Hawkes2017AGrowth}, and an encoder measures the spool’s rotation $\Delta \theta$ as it unwinds to release new material. The IMU at the robot tip outputs angular velocity $\Omega^{B}$, linear acceleration $A^{B}$, and magnetic field $M^{B}$, in its own body frame.  Applying filtering, data fusion, and a frame transformation algorithm, the linear acceleration $A^{G}$ is computed in the global frame, which is the same as the initial robot pose frame; and the robot tip velocity vector $V^{G}$ can be estimated by applying numeric integration over the sampling time. With real-time sensor information about current tip orientation $V^{G}$ and body length change $\Delta L$, the current position $P^{G}$ is estimated by summing the previous position with the instantaneous position change, as shown in the equation and notation in Fig.~\ref{fig:design_figure}(d).\par

\section{Experimental Characterization}
\subsection{Steering Device Reachable Workspace Characterization}

\begin{figure}[ht]
    \centering
    \includegraphics[width=9cm]{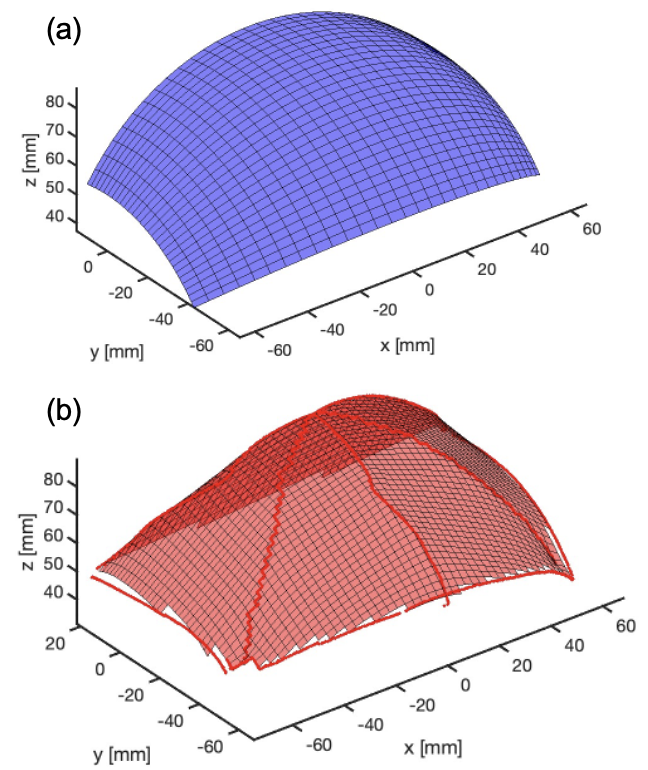}
    \vspace{-0.4cm}
    \caption{3D workspace analysis of the external steering device in frame ${2}$: (a) theoretical result showing a fitted truncated quarter sphere, and (b) experimental workspace mapped from observed tip positions (solid lines) with the same fitted surface overlaid (transparent mesh). Results are based on the characterization method described in Section IV.} 
    \vspace{-0.1cm}
    \label{fig:WS}
\end{figure}

To characterize the 3D workspace of the steering device, the device was positioned vertically (along the negative direction of gravity). A depth camera (Micron Tracker 3) was placed overhead to track the tip orientation by observing the pose of a marker frame attached to the tip of the steering device. The steering device was programmed to first sweep the workspace boundary and then traverse five intermediate lines: two lines with $\alpha = 0\degree$, one with $\theta = 0\degree$, one with $\theta = \alpha$, and another with $\theta = -\alpha$. In Fig.~\ref{fig:WS}(b), the solid red line represents the experimental data, while the transparent red surface indicates the corresponding interpolated surface. The experimental workspace resembled a truncated quarter-sphere surface, with the maximum steerable angles measured as $\alpha_{i,\text{max}} = 51.7\degree$ and $\theta_{i,\text{max}} = 56.0\degree$, and a sphere radius of 88.3 cm. The percentage errors for the maximum steerable angles were 1.54$\%$ and 6.67$\%$, respectively, and the percentage error for the effective radius was 2.17$\%$. The experimental measurements showed no significant deviation from the theoretical model.

\vspace{0.3cm}

\subsection{Blocked Force Measurements}

\begin{figure}[ht]
    \centering
    \includegraphics[width=8.7cm]{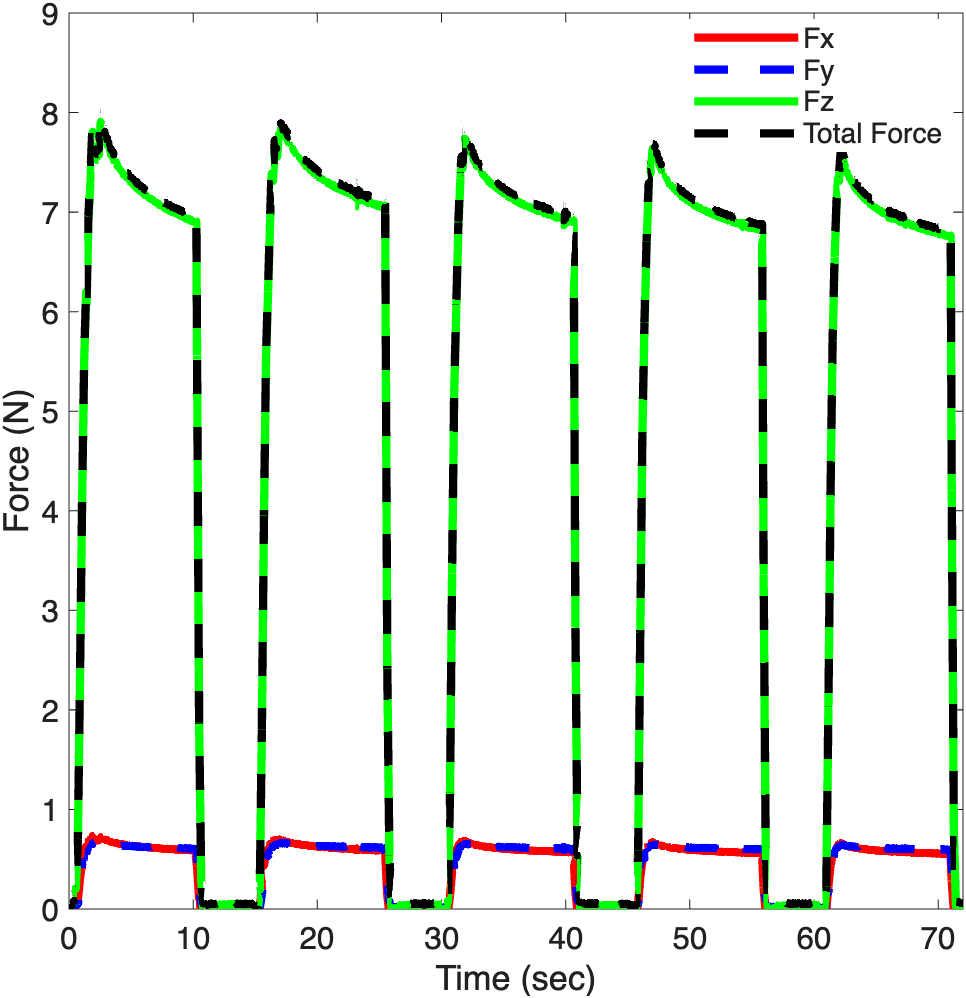}
    \caption{Blocked force measurement used to characterize the maximum tip load that the device is capable of maneuvering in 3D space: (red solid line) force in x direction, (blue dashed line) force in y direction, (green solid line) force in z direction, and (black dash line) total force.}
    \vspace{-0.2cm}
    \label{fig:blocked}
\end{figure}

\begin{figure*}[ht]
  \centering
  \includegraphics[width=\textwidth]{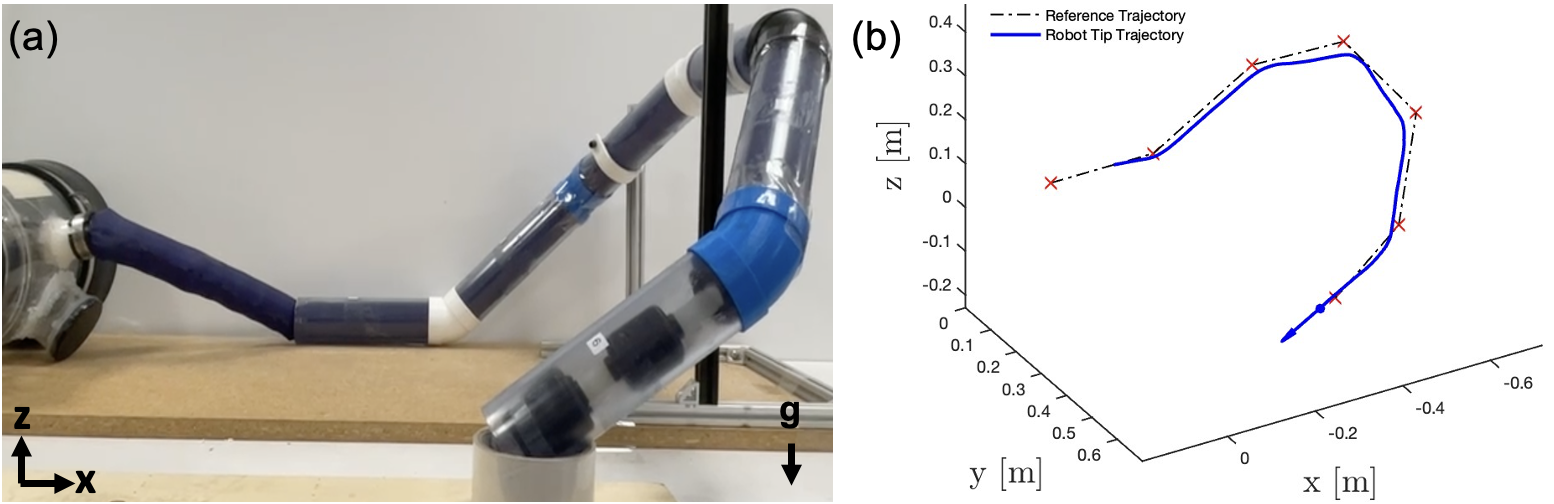}
  \caption{Robot tip localization algorithm relying on tip IMU sensors and encoders in the base station: (a) Robot passively traveling through a laboratory setup 3D pipe system consisting of five segments of 45$\degree$ turns. (b) The algorithm computes the tip trajectory (blue solid line) in real-time and compares it with the ground truth reference trajectory (black dashed line). }
  \label{fig:mapping}
\end{figure*}

\begin{figure*}[!t]
  \centering
  \includegraphics[width=\textwidth]{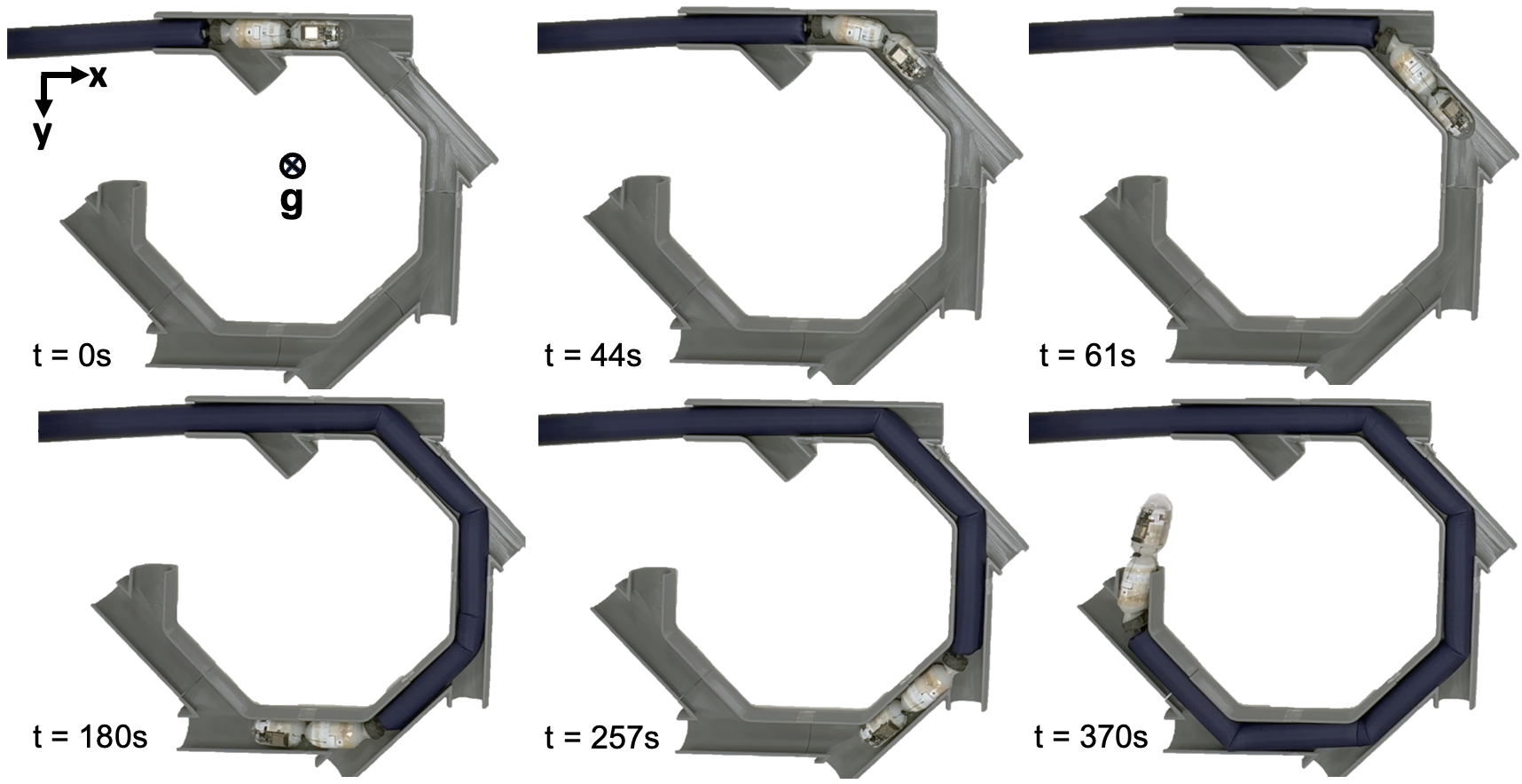}
  \caption{Robot tip steering movement over a customized 2D pipe system. The course is a series of 3D-printed pipe-like structures for the vine robot to progress through. Each piece has two possible directions for the robot: forward through passive movement and turning right at a 45$\degree$ turn. Seven of these pieces were assembled together and secured to the table for the demonstration. At t = 0s, the robot tip is passively moving due to the inflation of the "vine" robot. Then, for the rest of the pictures, the robot tip is actively navigating through the designed pipe maze.}
  \label{fig:xysteering}
\end{figure*}

\begin{figure*}[!t]
  \centering
  \includegraphics[width=\textwidth]{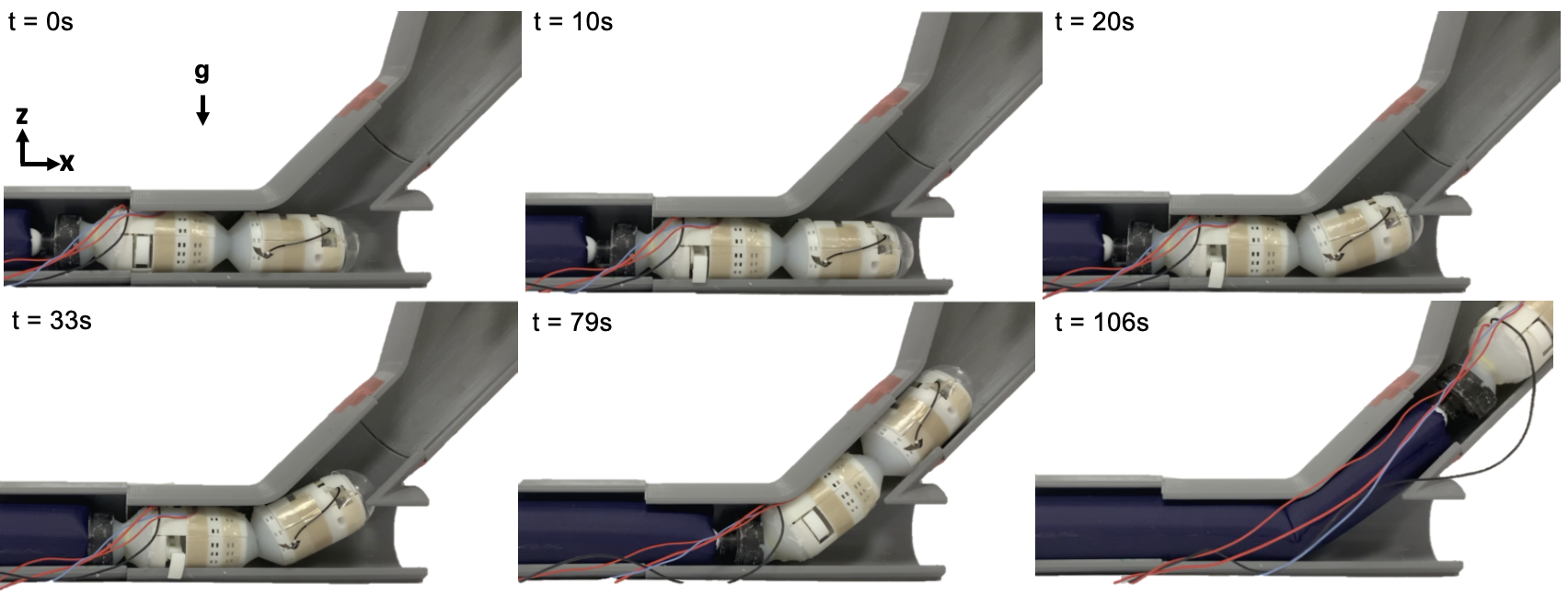}
  \caption{Robot tip steering movement over a pipe section with a 45° upward branch (opposing gravity direction).  To navigate this course, the robot first extends the bracing mechanism (t=10s), activates the tendon mechanism to orient up (t=20s), then propels forward due to the body inflation (t=79s), and finally retracts the bracing mechanism (t=79s). (Note: The visible power cables are only used for testing purposes; they are not required in the fully integrated system.)}

  \label{fig:verticalsteering}
\end{figure*}

Similar to the workspace characterization, the device was placed vertically (along the negative direction of gravity). A 6-axis force sensor (ATI Nano17 F/T Sensor) was mounted on the robot tip via a 3D-printed cap and connected to a rigid structure built from aluminum extrusions, which constrained the steering device in all directions to prevent movement or deformation of the silicone-molded joints during blocked force measurements. A microcontroller was incorporated to repeatedly send pulse width modulation (PWM) signals to the spool motor with period T = 15 sec and duty cycle 66.7$\%$. The PWM command was sent to the external steering tip several times to demonstrate the repeatability. As shown in Fig.~\ref{fig:blocked}, the device transmitted a peak force of magnitude 0.650 N, 0.640 N, and 7.40 N in the x, y, and z directions, respectively. These values were obtained by first measuring the raw forces and then compensating for gravitational effects—that is, subtracting the force contribution due to the weight of the robot tip from the sensor readings. Using Equation~\ref{eq:blockedforce}, we calculated an experimental motor stall torque at 3.3V operating voltage of 13.0 N-mm, which deviated by 3.20$\%$ error from the manufacturer data sheet. Fig.~\ref{fig:blocked} showed a decaying force in the z-direction over repeated trials. During the experiment, we observed that the blocked force gradually decreased, which we believed was due to mechanical degradation from the plastic gearbox used in the motor. Specifically, shaft misalignment and material wear in the plastic components may have led to increased internal friction and reduced transmission efficiency over time.

\section{Demonstrations}

\subsection{3D Pipe System Tip Localization}
To demonstrate the localization algorithm, described in Section~V, a customized 3D pipe system was set up in a laboratory environment with six segments of straight pipe (0.2m in length and 5cm in diameter). Adjacent segments were connected with a 3D-printed 45$\degree$ connector. As illustrated in Fig.~\ref{fig:mapping}(a), the vine robot carrying the sensor system passively navigated within the pipe system. The robot tip trajectory, computed by the proposed localization algorithm, was wirelessly provided to the operator in real-time as shown in Fig.~\ref{fig:mapping}(b). The black dashed line represented the center line of the pipe network, while the blue solid line was the calculated tip trajectory. Quantitatively, tracking error was defined as the error between the current robot tip location and the closest point on the reference trajectory. In this experiment trial, the tracking error was 180 mm, on average, with a standard deviation of 8.60 mm. During deployment, the main body pressure was controlled to be a constant internal pressure of 5.5 kPa via an electronic pressure regulator (Proportion Air QB3).


\subsection{2D Pipe Branching System Steering}
In this section, we showcase our design's capability for teleoperated steering in a constrained environment. A 3D-printed customized 2D branching pipe system, as shown in Fig.~\ref{fig:xysteering}, was designed with seven branching locations, where each consists of two possible directions: forward and 45\degree right turn. Every junction was connected by a straight pipe, which had a length of 15.0 cm and an internal diameter of 5.30 cm. During the deployment, the main body pressure was also controlled to be 5.5 kPa via a pressure regulator. The user selected branches through wireless teleoperation using keyboard input. Before entering a branch junction, we decreased the propulsion speed, sent a teleoperation command, and then steered toward the right branch. After the tip section passed the branch, as shown in Fig.~\ref{fig:xysteering} (t=44s), the steering tendon was released, allowing the soft joint to passively adapt to the environment geometry. At this moment, the robot body was propelled again, as shown in Fig.~\ref{fig:xysteering} (t=61s), which pushed the tip through the branch. This process was repeated for the remainder of the path. This demonstration showed the robot's maneuverability and control in branching pathways.

\begin{figure*}[!t]
  \centering
  \includegraphics[width=\textwidth,height=1.2\textheight,keepaspectratio]{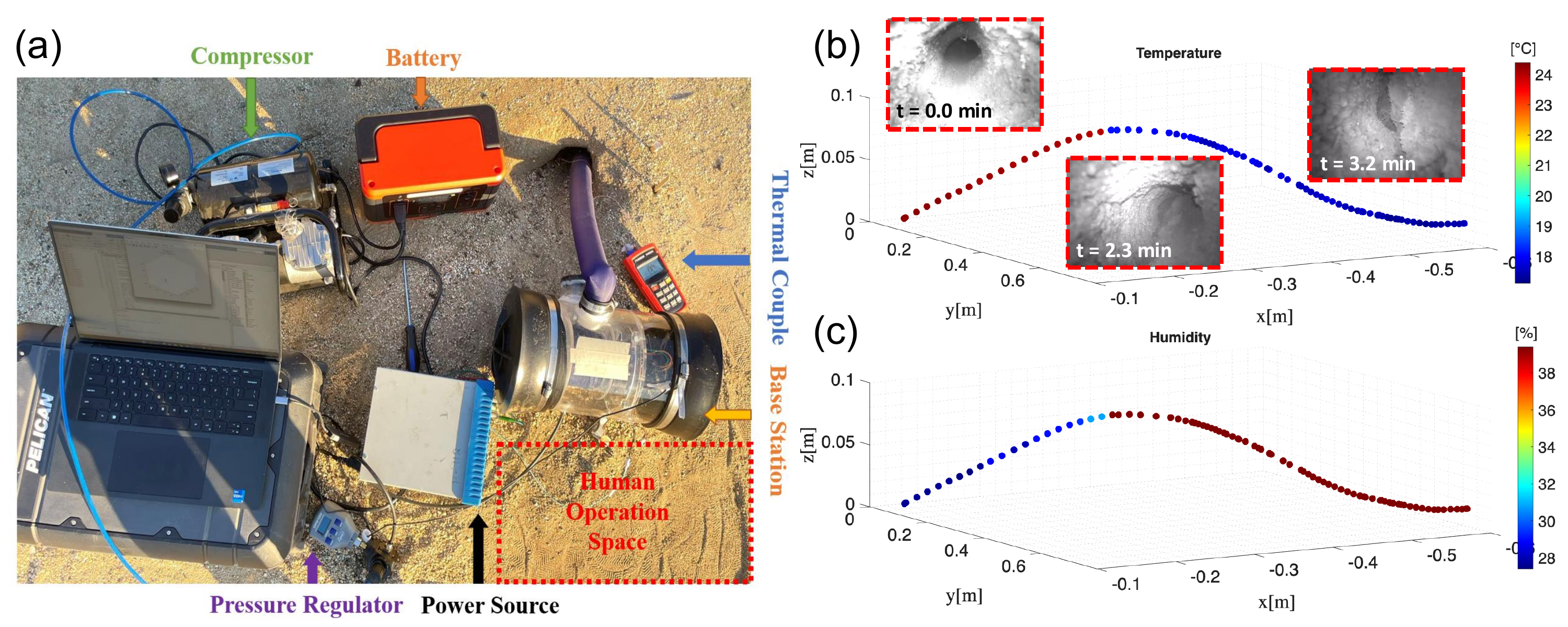}
  \caption{Tiger Salamander burrow investigation. (a) Field experiment setup for deploying the vine robot system in a tiger salamander burrow. (b) Reconstructed shape of a representative burrow with its temperature profile. Representative camera feedback is included in the figure. (c) Reconstructed shape of the same burrow with its humidity profile. Note: the temperature and humidity reading is not fully stabilized at the entry location.}
  \label{fig:field}
\end{figure*}

\subsection{Pipe Navigation Opposing Gravity}
To verify our design's steering ability in a 3D environment, we designed a pipe branching set that includes a straight pipe section (D=5.3cm) and a 45$\degree$ upward turning angle in the anti-gravity direction, as shown in Fig.~\ref{fig:verticalsteering}. The steering operation, in this part, was achieved by manually controlling the motor via an external DC power source. During the deployment, the robot was first propelled forward such that the tip of the steering device was at the center of the branch elbow (t=0s). The pressure was reduced to 0 kPa while waiting for the next phase to complete. The bracing mechanism was then expanded outward (t=10s), ensuring the robot remained centered in the straight pipe while providing stability through friction. Following the stabilization of the robot, the robot tip was steered in the vertical direction by shortening the top tendon cable (t=20s and t=33s). Then, the main body pressure was increased to 5.5 kPa to propel the vine robot forward (t=79s). After the tip section passed through the branch, the tendon and bracing mechanism were released to allow the robot body to passively navigate through the course.

\subsection{Tiger Salamander Burrow Investigation}

We demonstrated the practical capability of our system in a biological field study focused on understanding the habitat preferences of the California tiger salamander (CTS), an endangered subterranean amphibian species native to California \cite{cts_gov}. The CTS has faced significant habitat loss due to urbanization, which has severely impacted its ability to find suitable environments for survival \cite{cts_gov}\cite{cts_urban}\cite{cts_review}. While much is known about its terrestrial behavior, such as migration patterns and breeding, little is understood about its underground activity. The complex, small-diameter burrow systems used by CTS present challenges for conventional underground navigation and study methods.

Our field tests aimed to explore the tortuous burrow systems using the robotic system described in this letter, reconstruct burrow geometry, and collect environmental data such as temperature and humidity through our tip-sensing system. These tests were conducted at a known CTS habitat (location anonymized for double-anonymous review)\cite{cts_stanford}. A representative burrow investigation (diameter $\approx$ 5 cm) is shown in Fig.~\ref{fig:field}. The experiment setup included a pneumatic compressor, pressure regulator, portable battery, power supply, and the vine robot system, as depicted in Fig.~\ref{fig:field}(a). The described system was teleoperated within the burrow using real-time visual feedback from the tip-mounted camera. During the trial, the wireless signal maintained a clear connection with the human operator. During in-burrow navigation, the flexible, external tip steering design enabled the robot to adapt to channels with varying diameters. Additionally, the vine robot system generated sufficient propulsive force to overcome obstructions such as sand and small rocks. The robot successfully traversed the burrow until reaching a dead end. The burrow geometry was reconstructed using a localization algorithm, as shown in Figs.~\ref{fig:field}(b) and \ref{fig:field}(c). The trajectory indicates that the burrow extended 0.073 m vertically in the anti-gravity direction, with displacements of -0.425 m and 0.797 m in the x and y directions, respectively. This corresponds to a continuous bending angle of 62.0$^\circ$ in the x-y plane.

In addition to the visual feedback and localization sensing, a temperature and humidity sensor (HiLetgo DHT22) was integrated into the system to collect environmental data, with a resolution of 0.1$^\circ$C and 0.1$\%$ relative humidity. The collected data revealed that the salamander’s burrow maintains a temperature of approximately 17.2$^\circ$C and a humidity of 39.4$\%$. These results demonstrate the potential of our tip-sensing system to measure the environmental characteristics of CTS habitats.

\section{Conclusion and Future Work}
We present the design of an external tip steering device for constrained environment exploration and navigation, which addresses the limitations of previous vine robot systems. This design enables active 3D steering with large local curvature, allowing precise maneuverability in confined spaces with a small radius of curvature. Additionally, it includes a compliant tip design, enabling flexibility to navigate sharp turns. The device supports real-time 3D tip localization in GPS-denied environments. We present kinematic models and experiments to demonstrate the enhanced workspace achieved with the design, along with force analysis to evaluate its payload capacity for carrying sensors. We validate the system's feasibility through tests conducted in laboratory 3D pipe systems and natural tiger salamander burrows.

The steering methods demonstrate potential for inspection in confined environments. The system integrates a miniaturized external steering mechanism and a compliant robot body, allowing traversal through complex geometries like machinery interiors, pipe systems, and animal burrows. Tip-mounted sensors support environmental data collection and localization of the robot tip. For example, the system can be deployed to detect corrosion in tortuous pipelines and enclosed tanks.

Despite the system's demonstrated capabilities, several limitations remain. The use of rigid components, such as motors, 3D-printed housings, and electronic boards, inherently limits the minimum diameter of navigable environments and the achievable bending radius. Moreover, excessive tail tension in tortuous environments can significantly restrict the deployable length of the vine robot, suggesting the importance of decreasing tail tension to enable long-distance deployment. Additionally, incorporating visual data into current localization algorithms could improve accuracy by compensating for drift in accelerometer and angular velocity measurements. Finally, data communication can be improved by integrating a long-distance tethered transmission method, such as Ethernet, to enable data transfer in wireless-signal denied environments.

\section{Acknowledgments}
The authors thank Frances Raphael for his assistance with the field experiments, and Dr. Esther Marika Cole Adelsheim, Conservation Program Manager, for providing access to the field site and for her insightful discussions on animal biology.

\newpage
\bibliographystyle{IEEEtran}
\bibliography{references1}

\begin{thebibliography}{10}
\bibitem{cts_stanford}
``{California tiger Salamander (Ambystoma californiense)},'' [URL omitted for
  anonymous review], details will be included in the final version.
\end{thebibliography}

\ifCLASSOPTIONcaptionsoff
  \newpage
\fi




%

\end{document}